\RequirePackage{fix-cm}

\documentclass[smallextended]{svjour3}       
\smartqed  
\usepackage{graphicx}
\usepackage{multirow}
\usepackage{amssymb}
\usepackage{amsmath,xparse}
\usepackage{footnote}
\makesavenoteenv{table}
\usepackage[misc]{ifsym}
\usepackage{CJKutf8}  
\usepackage{pinyin}  
\usepackage{threeparttable}
\usepackage{soul,color}
\sethlcolor{yellow} 

\title{Fine-Grained Emotion Classification of Chinese Microblogs Based on Graph Convolution Networks
\footnote{The first two authors contributed equally to this paper.}}
\titlerunning{Emotion Classification of Chinese Microblogs Based on GCN}        

\author{Yuni Lai$^1$\thanks{$^1$Northeastern University, Shenyang, 110819, China}\and Linfeng Zhang$^1$\and\Letter Donghong Han$^1$
	\and Rui Zhou$^2$\thanks{$^2$Swinburne University of Technology, Hawthorn, 3122, Australia}
    \and Guoren Wang$^3$\thanks{$^3$Beijing Institute of Technology, Beijing, 100081, China}}

\begin{document}
\institute{%
\Letter Donghong Han 
	\at{email:handonghong@cse.neu.edu.cn}	
}	
\date{Received: date / Accepted: date}

\maketitle
\begin{abstract}
Microblogs are widely used to express people’s opinions and feelings in daily life. Sentiment analysis (SA) can timely detect personal sentiment polarities through analyzing text. Deep learning approaches have been broadly used in SA but still have not fully exploited syntax information. In this paper, we propose a syntax-based graph convolution network (GCN) model to enhance the understanding of diverse grammatical structures of Chinese microblogs. In addition, a pooling method based on percentile is proposed to improve the accuracy of the model. In experiments, for Chinese microblogs emotion classification categories including happiness, sadness, like, anger, disgust, fear, and surprise, the F-measure of our model reaches 82.32\% and exceeds the state-of-the-art algorithm by 5.90\%. The experimental results show that our model can effectively utilize the information of dependency parsing to improve the performance of emotion detection. What is more, we annotate a new dataset for Chinese emotion classification, which is open to other researchers.
\keywords{Sentiment Analysis \and Graph Convolution Network \and Chinese Microblog \and Deep Learning \and Emotion Detection}
\end{abstract}

\clearpage\section{Introduction}
\label{intro}
Sentiment analysis (SA) is a problem belonging to natural language processing (NLP) to detect sentiment polarities of a writer from a piece of text. In the research field of SA, fine-grained emotion classification aims to detect exact types of feelings such as \emph{happiness, sadness, like, anger, disgust, fear,} and \emph{surprise}. Nowadays, more and more Chinese people share their daily feelings on social networks. For example, one of the most popular Chinese microblog platforms, Sina microblog, has its monthly active users increased to 431 million by August 2018 \cite{00}. SA for social network has become a hot topic in recent years, and the analysis results can be widely used in public opinion analysis, psychological research, social events and even political elections \cite{Ref1}.
\par 
Sentiment analysis (or emotion detection) for English tweets is quite satisfactory in the SemEval-2017 with 48 teams participated \cite{Ref2}. However, there are not as much research in the field of Chinese. One of the key issues in emotion analysis for Chinese tweets is how to understand text with various and complex syntactic structures. When a person is reading a sentence, she reads words one by one and has grammar structures in mind to comprehend the entire sentence. Guided by this reading behaviour, a deep learning model of long short-term memory (LSTM) \cite{Ref3} was proposed to deal with NLP problems. However, although LSTM has a long memory of essential words, it may have not fully exploited the syntactic information of text to see the sentence as a whole as a person.  
\par
\begin{figure}[htb]
	\includegraphics[width=8cm]{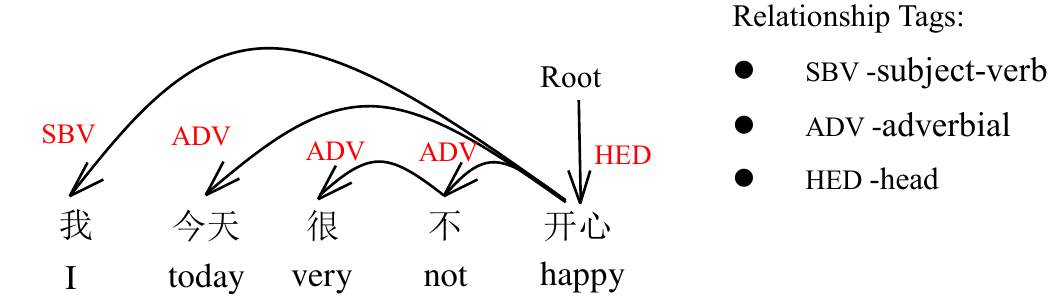}
	\caption{An example of dependency parse tree}
	\label{fig:Fig.1}       
\end{figure}

\begin{CJK}{UTF8}{gbsn} 
In fact, syntax structure is important for a model to identify the emotion of texts, especially for microblogs that contain the same set of words but different syntaxes. For example, the meanings of the two sentences "Don't you know that?" and "You don't know that!" are different since the former one expresses 'surprise' while the latter one is probably more negative. 
Motivated by this, we consider dependency parse tree, which reveals the syntactic dependency relations among language components, could be of help. Besides, according to other existing research, dependency parse tree did help to increase the performance of other NLP tasks like event detection {\cite{Ref35}} and semantic role labeling {\cite{Ref36}}. Specifically, syntactic dependency identifies subject-verb, adverbial and other grammatical structures and then analyzes the relationship between different elements in a sentence. Fig.\ref{fig:Fig.1}  shows the dependency analysis result of an example sentence (‘\footnotesize{我今天很不开心}’;\normalsize{ I am very unhappy today). If there is an edge between two words, these words are related \cite{Ref4,Ref5}.}
\end{CJK} 
\par
Dependency parse tree is a kind of non-Euclidean structure, which cannot be processed by convolution neural networks (CNN) \cite{Ref6}, a widely-used deep learning network. In order to apply deep learning on non-Euclidean domain, graph convolution network (GCN) \cite{Ref31} is proposed to extend convolution to graph-structure domains. As a result, in order to integrate grammar information and better simulate the reading process of human, we propose a syntax-based GCN model to detect emotion polarities. Firstly, bidirectional long short-term memory network (Bi-LSTM) \cite{Ref7} is adopted to extract preliminary features for each word. Then these features and dependency parse tree are fed in GCN. Dependency parse tree based on text is used to build the convolution graph for GCN. In the graph, each word is regarded as a vertex, and each syntactic dependency relation between two words is regarded as an edge.
In the operation of GCN, each word feature given by Bi-LSTM, regarded as the feature of each vertex, is convoluted according to the graph. GCN obtains a final representation vector of each microblog for classification. The convolution over dependency parse graph can focus on relevant words to promote the ability to capture complex linguistic phenomena.
\par
In this paper, we tackle the challenge of Chinese fine-grained emotion classification, including seven types of basic emotions---\emph{happiness, sadness, like, anger, disgust, fear,} and \emph{surprise}. To the best of our knowledge, this is the first work to integrate dependency parse tree in the graph convolution network model for emotion detection. Our main contributions of this paper are as follows:
\par 
\begin{itemize}
	\item[$\bullet$] We propose a novel syntax-based graph convolution network (GCN) model for emotion detection of Chinese microblogs. After preprocessing, a set of embedding vectors representing a microblog are fed into a Bi-LSTM neural network. Then, GCN applies convolution on the preliminary word features from Bi-LSTM according to syntactic trees, extracting comprehensive features of microblogs.
	\item[$\bullet$] In order to improve the performance of the model, we propose a pooling method of percentile pooling, which enhances the robustness of the model since percentile is not sensitive to outliers. 
	\item[$\bullet$] Experimental results show that the proposed model can fully exploit the syntactic relation of a sentence and outperform state-of-the-art algorithms. In addition, we collected and labeled 15,664 microblogs to construct a high-quality Chinese microblogs dataset which is open to other researchers. Our experiment codes and datasets are free on GitHub\footnote{https://github.com/zhanglinfeng1997/Sentiment-Analysis-via-GCN}.
\end{itemize}
\par 
The rest of paper is organized as follows: In Section 2, we introduce the background and related work. Our model and novel algorithms are presented in Section 3, with experimental results and evaluations discussed in Section 4. Finally, Sections 5 gives the conclusions and future work.

\section{Related Work}
\subsection{Sentiment Classification}
Sentiment analysis (SA) tasks have attracted much attention these days, including two main types: coarse-grained sentiment polarity classification and fine-grained emotion classification. Our work belongs to the latter one. 
\par 
In recent years, deep learning methods are applied to analyze text sentiment, such as convolution neural network (CNN) and recurrent neural network (RNN). Yoon Kim \cite{Ref9} used CNN trained on top of pre-trained word vectors for sentence-level classification tasks. After that, much research has been done to improve the CNN models as well as word representation. Kalchbrenner et al. \cite{Ref10} described a dynamic convolution neural network with dynamic k-max pooling, modeling the semantics of sentences. Chen et al. \cite{Ref11} built a character embedding with dual channel convolution neural network to comprehend the sentiment of Chinese short comments in Sina Weibo. Their model reached a relatively high accuracy of 88.35\% on the NLP\&CC2012 dataset for binary classification. Zhao et al. \cite{Ref12} introduced a word embedding method combining n-grams features and word sentiment polarity score features. They integrated the feature set into a deep convolution neural network to train the sentiment classifier. Xue \& Li \cite{Ref13} proposed a model based on CNN with gating mechanisms, which had higher accuracy and efficiency. Unfortunately, these CNN-based works have not made full use of syntax information of text. 
\par 
Other researchers used extended recurrent neural network (RNN) to classify sentiment polarity. Socher et al. \cite{Ref14} introduced a sentiment treebank and recursive neural tensor network which outperforms all previous methods on several metrics. Their work employed syntactic information on Stanford Sentiment Treebank corpus with full sentiment labeled parse trees in neural networks.  In order to design a document-level sentiment classification model, Tang et al. \cite{Ref15} employed LSTM to extract the sentence representation and then used gated recurrent neural network to encode document representation. Baziotis et al. \cite{Ref16} used deep LSTM along with context-aware attention mechanisms to conduct topic-based sentiment polarity analysis in Twitter. Similarly, Ma et al. \cite{Ref17} integrated commonsense knowledge into LSTM model extended with target-level attention and sentence-level attention. Although RNN is a suitable network for NLP, if it is used alone, its learning ability becomes limited when the task getting complex, because during the operation, RNN throws away some information that may be useful in some situation and fails to give an overall view. Instead of applying RNN alone, we combined LSTM with graph convolution network, which complements the shortcomings of LSTM while keeps its advantages. 
\par 
Instead of using CNN or RNN alone, some researchers combined two networks together to obtain better performance. Wang et al. \cite{Ref18} proposed a CNN-LSTM model including regional CNN and LSTM to gain the valence-arousal ratings of texts, local and overall information. 
Our idea is similar to it, stacking two different neural networks to improve the performance of the deep learning model. However, none of the existing works have combined LSTM with syntactic graph convolution network for sentiment analysis as we do in this paper. 
\par 
Besides improving network structures, some research tried to improve models with linguistic information. Qian et al. \cite{Ref19} attempted to model the semantic role of sentiment lexicons, negation words, and intensity words in sentiment expression. Meanwhile, they expressed these linguistic factors with mathematical operations, parameterized with shifting distribution vectors or transformation matrices. Similarly, Lei et al. \cite{Ref20} integrated three kinds of sentiment linguistic knowledge (e.g., sentiment lexicon, negation words, intensity words) into the deep neural network via attention mechanism and proposed a multi-sentiment-resource enhanced attention network to better exploit sentiment linguistic knowledge. Although they also integrated linguistic content to train models, sometimes there is no sentiment lexicon information in some microblogs. These works did not used dependency parse tree to train a deep learning network. Since dependency parse tree is a quite mature technique that has been successfully used in language understanding, it can reveal syntax structure of sentences more accurately and clearly. 
\parskip=0pt\par 
There are more research efforts focusing on fine-grained emotion detection. Some works are based on traditional machine learning methods, rule-based, emotion lexicon, and emoticons. Li \& Xu \cite{Ref21} designed a support vector regression method along with rule-based emotion caused extraction to classify 6-point emotion for Sina Weibo. Wen \& Wan \cite{Ref22} introduced an approach based on sequential class rules for emotion classification of microblog, in which each tweet was regarded as a sequence. They used emotion lexicons and machine learning to obtain potential emotion labels in tweets. Zheng \& Purver \cite{Ref23} described a method for detecting emotions of Chinese microblog using distant supervision with conventional labels (emoticons and smileys). Unfortunately, the process of generalizing and designing linguistic patterns for posts with complicated linguistic patterns is labor-intensive and time–consuming. Different from rule-based methods above, we design an automatic model without human generated rules.
\par 
Recently, deep neural networks were employed for emotion detection. Wang et al. \cite{Ref24} leveraged the skip-gram language model to learn distributed word representations as input features and utilized a CNN-based method to classify multi-label emotion sentences of Chinese microblogs. Muhammad \& Lyle \cite{Ref25} built an enormous, automatically curated dataset for emotion detection using distant supervision and then modeled with gated recurrent neural networks, achieving an average accuracy of 95.68\% in 8-point classification on English text. He et al. \cite{Ref26} proposed an emotion-semantics enhanced multi-channel convolution neural network model that used emoticons to construct emotion space, combining different embedding methods to extract features which would be fed into multi-channel convolution neural network. He \& Xia \cite{Ref3} proposed a joint Bi-LSTM network model to detect multi-label emotion on a Chinese long blog corpus. Different from works above, we integrate dependency parse tree that can be obtained by existing algorithms to make our model more suitable for fine-grained task, aiming to enhance deeper understanding of text by introducing grammar information to the model.
\par 
Although sentiment analysis has been developed for several years, the emotion classifying accuracy for Chinese microblogs is still not satisfactory due to the complexity of Chinese. In this paper, we focus on this issue and adopt Bi-LSTM model and graph convolution neural network to enhance emotion understanding of microblogs.

\subsection{Pooling Method}
Pooling layer plays an important role in image processing and deep learning since it helps to keep values we need and discard others. 
In image quality assessment field, percentile pooling has been introduced to detect the poor-quality regions \cite{Ref27,Ref28}. According to a common experience that deep neural network is expected to obtain a higher value if the input is close to the learned pattern, max pooling is widely-used to extract the maximum value for further analysis. Max pooling, as a special case of percentile pooling, performs well in neural networks, but there is still room for improvement. 

Based on max pooling, k-max pooling, keeping $k$ top maximum values in order to include more information, was designed by Kalchbrenner et al, which has better performance in some cases \cite{Ref29}. 

Another popular global pooling method is average pooling \cite{Ref30} that keeps the mean value of a group.

Different from the works above, we introduce percentile pooling into neural networks where the percentile is not restricted to the maximum since the specific percentile is adjustable.


\subsection{Graph Convolution Neural Network}

There are lots of graph structure data in the world such as social network among people, web page link network and paper citation network. In these applications, every node object not only has its own feature but also has its connection information to others. For example, in the social network, each person has his/her feature, such as age, hobbies and job. Besides, each person makes friends with different people. If we want to classify these people, using the connection information is necessary. However, CNN can only take the feature information. In order to take the connection information as well, Kipf \& Welling \cite{Ref31} proposed a spectral graph convolution network (GCN) that uses localized first-order approximation to encode both local graph structure and features of nodes, which speeds up the process of training and works well in semi-supervised classification. After that, GCN draws more attention of researchers and then was employed in a variety of applications such as web-scale recommendation systems \cite{Ref32}, Skeleton-based action recognition \cite{Ref33}, and traffic forecasting \cite{Ref34}. In NLP tasks, GCN also has good performance in event detection \cite{Ref35}, semantic role labeling \cite{Ref36}. Considering the significance of syntax structure in sentiment analysis and the graph-structure of dependency parse tree, we propose to use GCN to integrate grammar tree in order to better extract emotions of complex Chinese text. To the best of our knowledge, this is the first work to employ GCN in Chinese emotion detection.

\section{Syntax-based Graph Convolution Neural Network Model}
Most of the microblogs have a variety of emotion tendencies while some texts do not express any sentiment. In practical applications, such as psychological research or user emotional portrait, emotion analysis for social networks is necessary. According to the task of NLP\&CC2013, in this paper, we focus on the research of classifying Chinese microblogs with sentiment into seven categories: \emph{happiness, sadness, like, anger, disgust, fear,} and \emph{surprise}.
\par 
The model is mainly composed of three parts: (i) First, a Bi-LSTM network is used to extract preliminary word features of the given text. (ii) Then, we feed the preliminary word features and the dependency parse tree built for each microblog into a single-layer graph convolution network (GCN) to exploit the emotion feature of the microblog. (iii) Finally, we obtain the probability distribution with pooling or fully connected layer. As Fig.2 shows, our model has the following steps:

\begin{itemize}
\item[$\bullet$] Preprocessing
\item[$\bullet$] Word embedding 
\item[$\bullet$] Bi-LSTM layer
\item[$\bullet$] GCN layer
\item[$\bullet$] Pooling or fully connected layer
\item[$\bullet$] Softmax classification
\end{itemize}

\begin{figure}
	\includegraphics{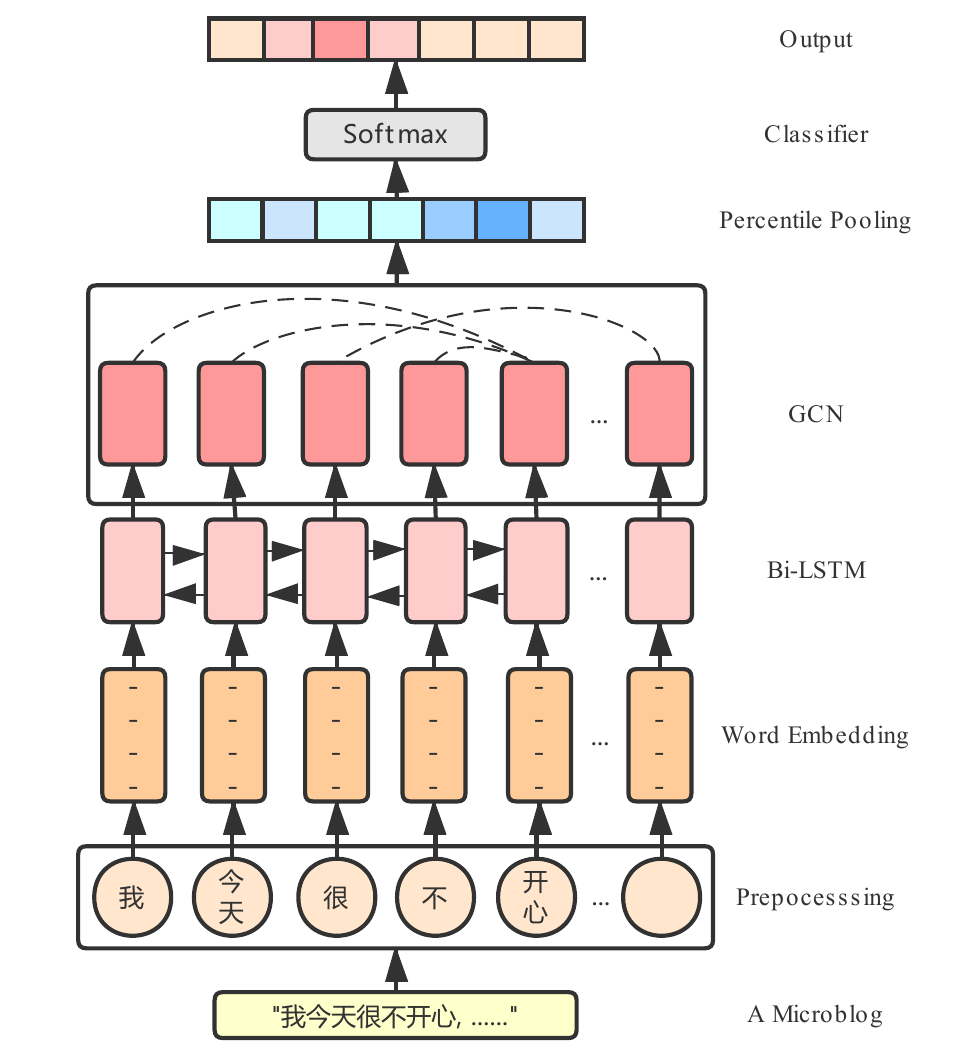}
	\caption{Syntax-based GCN model structure}
	\label{fig:2}       
\end{figure}

\subsection{Preprocessing}
Raw microblog data usually have redundancy and noise such as the URLs, ‘@’ symbols and useless stop words. We first clean up all the unnecessary contents of the text. Next, jieba\footnote{https://github.com/fxsjy/jieba}  python package is adopted to realize Chinese sentence segmentation, separating the sentences into words. Then, we use the ltp\footnote{http://ltp.ai}  python package to get the dependency parse tree of each microblog. For word embeddings, we use random initialization method to represent each word by a random vector, and the vector will be updated during network training.
\par 
We denote each microblog after preprocessing as $X=\left\{ x_{1},x_{2},\ldots,x_{n}\right\}$, in which $ x_{i}\in \mathbb{R}^{300} \left( i=1,\ldots, n\right)$  is the embedding vector representing a word. Then $X$ would be fed into a Bi-LSTM neural network.
\subsection{Bidirectional LSTM Neural Network}
A Bi-LSTM is used to obtain the rudimentary representations of microblogs. After applying Bi-LSTM on word embedding sequence $X=\left\{ x_{1},x_{2},\ldots,x_{n}\right\}$ of a microblog, we can get both forward and backward vector sequences, denoted by $L_{1}=\left\{ l_{11},l_{12},\ldots,l_{1n}\right\}$ and $L_{2}=\left\{ l_{21},l_{22},\ldots,l_{2n}\right\}$ respectively. We concatenate the two sequences to get $L=\begin{bmatrix} L_{1} \\ L_{2} \end{bmatrix}=\left\{ l_{1},l_{2},\ldots,l_{n}\right\}$, where $l_{i}=\left[l_{1i},l_{2i}\right]^\mathrm{T}, i=1, \ldots,n$. $L$ is the elementary feature of $X$. Here, each word in a sentence has its own word feature $l_i$. Next, each output vector $l_i$ from Bi-LSTM and the syntactic trees will be the input of the GCN network. 
\subsection{Graph Convolution Network}
\begin{CJK}{UTF8}{gbsn} 
For each microblog, a graph $G=\left(V,E\right)$ is built, where $V$ is the vertex set consisting of all words of a microblog, and $E$ is the edge set including all dependency relations between two words. According to Kipf \& Welling \cite{Ref31} and Marcheggiani \& Titov \cite{Ref36}, we add self-loops and opposite edges to the edge set, which can improve the general ability of GCN. The numbers `0', `1', `1', `1' are applied to label the types of dependency relations of no relation, self-loop relation, head-to-dependent, and dependent-to-head, respectively. Based on these rules, the sparse adjacency matrix of the dependency parse tree, denoted by $A$, for each blog is created. Fig.3 shows the corresponding adjacency matrix for the example sentence (`\footnotesize{我今天很不开心}';\normalsize{ I am very unhappy today) according to its dependency parse tree. For words of different sentences in the same microblog, we regard no relation between them, and the labels of the edges are all `0'. Since every microblog is no longer than 140 words, we set every adjacency matrix $A$ of size $\left[ 140\times140\right]$, and padding with `0'.}
\end{CJK}
\begin{figure}
	\includegraphics[width=10cm]{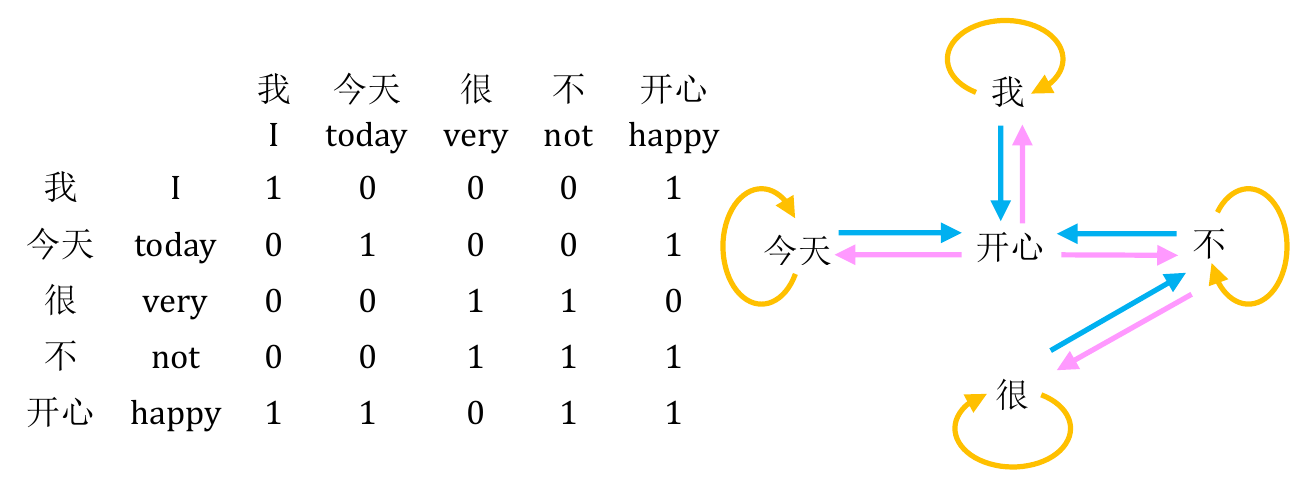}
	\caption{The adjacency matrix of a sentence and its relationship}
	\label{fig:3}       
\end{figure}
\par 
There is one-to-one correspondence between the nodes and word feature vectors $\left\{ l_{1},l_{2},\ldots,l_{n}\right\}$ given by Bi-LSTM, and each $l_i, \left( i=1,\ldots, n\right) $  is the word feature of each vertex. GCN applies convolution on features of the vertices according to the graph $G$ which is represented by the adjacency matrix A. 
\par According to Kipf \& Welling \cite{Ref31}, GCN network can be calculated as:
\begin{equation}
 Z=ReLU\left( \widetilde{D}^{-\frac {1}{2}}A\widetilde{D}^{-\frac {1}{2}}L\Theta\right),
\end{equation}
where ReLU is the rectifier linear unit activation function; $\widetilde{D}$ is the degree matrix of dependency tree, which is attained by $ \widetilde{D}_{ii}=\sum_{j}A_{ij}$; $A$ is a sparse adjacency matrix of a dependency parse tree; $L=\left\{ l_{1},l_{2},\ldots,l_{n}\right\}$ is a matrix that made of vectors given by Bi-LSTM; $\Theta$ is the weight matrix that learned by the network training.

\subsection{Percentile Pooling}
The aim of the pooling layer is to improve the invariance and the efficiency of the neural network model. In general, max pooling performs better than average pooling when dealing with NLP problems \cite{Ref37}. However, sometimes max pooling is not robust enough since there may be some uncontrollable and unexpected noise in models, which would cause the maximum of a group of numbers extremely high. To solve the problem, we adopted a novel pooling method based on percentile, named $p$th percentile pooling. $p$th percentile means the lowest $p$\% value of a set when the elements have been ascendingly sorted \cite{Moorthy}. As we know, $p$th percentile like the 50th percentile (median) is more robust than average since it is not sensitive to outliers. Besides, percentile pooling also have its advantage compared to max pooling. Max pooling is a special case of percentile pooling. However, the maximum value might not be the best choice all the time. If there is a very large outlier in a group of numbers, max pooling always output the outlier. In this case, 90th percentile pooling can be of help to better represent the top 10\% large values while excluding the outlier as shown in Fig. 4. We denote the $p$th percentile for a vector $Z$ as a function $f_{p}(z)$, in which the value $p$ ranges from 0 to 100. For example, $f_{100}(z)$  is the maximum and  $f_{50}(z)$ is the median of $Z$. We use $f_{p}(z)$ as the pooling function for $p$th percentile pooling. Here, $p$ is a hyper-parameter that can be adjusted until the best result is  obtained, which provides researchers with a more flexible choice aside from max pooling.
\begin{figure}
\includegraphics[width=10cm]{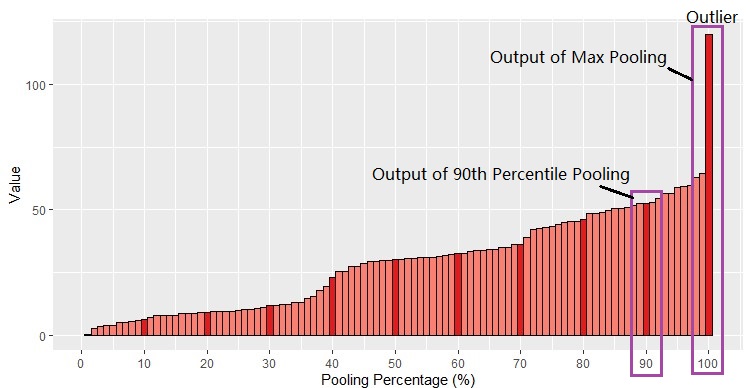}
\caption{The advantage of percentile pooling compared to max pooling.}
\label{fig:44}       
\end{figure}

\subsection{Orthogonalization Constraint}
In order to control the problem of vanishing and exploding gradients, we employ an orthogonalization constraint in training. A convenient way to achieve an approximate orthogonalization constraint is to add a regularization term in the loss function as follow,
\begin{equation}
Loss=loss\left( y,f_{w}\left( x\right) \right)+\lambda \sum _{i}\left\| W^{T}_{i}W_{i}-I\right\| ^{2},
\end{equation}
where $loss()$ is the original loss function, $y$ is the label, and $f_{w}$ is the predicted class, $\lambda$ is the penalty coefficient, $W_{i}$ is the weight matrix, and $I$ is an identity matrix. Besides, weight matrices in Bi-LSTM and GCN are initialized with orthogonal matrices. We employ singular value decomposition (SVD) on a randomly initialized matrix $M$, and we get $M=USV^{T}$, where $U$ and $V$ are orthogonal matrices and $S$ is a diagonal spectral matrix. $U$ or $V$ can be used to initialize weight matrix $W$, that is $W:= U$. 
\section{Experiments and Results Analysis}
All the codes are implemented on PyTorch 0.4.0 running on Linux CUDA platform. 
It takes around 2 hours to train a syntax-based graph convolution network (GCN) model on a 1080Ti GPU computer.
\subsection{Datasets}
In our experiments, the NLP\&CC2013 dataset\footnote{http://tcci.ccf.org.cn/conference/2013/dldoc/evdata02.zip} is chosen, which was used for sentiment classification task for the International Conference on Natural Language Processing and Chinese Computing (NLP\&CC). To improve the generalization ability of our model, we crawled 15,664 microblogs from Sina Weibo randomly and then labeled them by three human judges. Inconsistent labels were determined by votes. The test dataset is provided by NLP\&CC2013 for testing, and all the remaining are used for training. Table 1 shows the distribution of our datasets.

\begin{table}[htb]
	\caption{Datasets}
	\label{tab:1}
	\begin{tabular}{cccc}
		\hline
		\multirow{3}{*}{Emotion types} & \multicolumn{2}{c}{Number (training)} & Number (testing) \\ \cline{2-4} 
		& Training set of & Randomly crawled & Testing set of \\
		& NLP\&CC2013 & (Self-annotated) & NLP\&CC2013 \\ \hline
		Happiness & 940 & 2,797 & 370 \\
		Sadness & 633 & 2,467 & 385 \\
		Like & 1,284 & 4,257 & 595 \\
		Anger & 360 & 1,896 & 235 \\
		Disgust & 814 & 3,129 & 425 \\
		Fear & 96 & 299 & 49 \\
		Surprise & 211 & 819 & 113 \\ \hline
		Total & 4,338 & 15,664 & 2,172 \\ \hline
	\end{tabular}
\end{table}

\subsection{Performance Measure}
In this paper, since the testing dataset is originated from NLP\&CC Emotion Analysis in Chinese Weibo Text task\footnote{http://tcci.ccf.org.cn/conference/2013/dldoc/ev02.pdf} , we use the same metrics as before for the convenience of comparison. The evaluation metrics are macro average and micro average on precision, recall, and F-measure which defined as follows:
\begin{equation}
Macro_{Percision}=\frac{1}{7}\sum_{i}\frac{\#system\_correct\left(emotion=i\right)}{\#system\_proposed\left(emotion=i\right)},
\end{equation}
\begin{equation}
Macro_{Recall}=\frac{1}{7}\sum_{i}\frac{\#system\_correct\left(emotion=i\right)}{\#gold\left(emotion=i\right)},
\end{equation}
\begin{equation}
Macro_{F\_measure}=\frac{2\times Macro_{Percision}\times
	Macro_{Recall}}{Macro_{Percision}+Macro_{Recall}},
\end{equation}
\begin{equation}
Micro_{Percision}=\frac{\sum_{i}\#system\_correct\left(emotion=i\right)}{\sum_{i}\#system\_proposed\left(emotion=i\right)},
\end{equation}
\begin{equation}
Micro_{Percision}=\frac{\sum_{i}\#system\_correct\left(emotion=i\right)}{\sum_{i}\#system\_proposed\left(emotion=i\right)},
\end{equation}
\begin{equation}
Micro_{F\_measure}=\frac{2\times Micro_{Percision}\times
	Micro_{Recall}}{Micro_{Percision}+Micro_{Recall}},
\end{equation}
where $\#gold$ is the number of labels manually annotated for the test set, $\#system\_proposed$ is the number of classified tags by our system on the test set, and $\#system\_correct$ is the number of microblogs which are correctly classified. $i$ is one of the emotion type among \emph{happiness, sadness, like, anger, disgust, fear,} and \emph{surprise}.
\subsection{Hyper-parameters and Training}
The details of hyper-parameter setting in syntax-based graph convolution network (GCN) model is shown in Table 2. 
\begin{table}[htb]
	\caption{Hyper-parameters setting}
	\label{tab:2}
	\begin{tabular}{ccc}
		\hline
		& Parameters & Values / Descriptions \\ \hline
		Word embedding & Embedding size & 300 \\ \hline
		\multirow{5}{*}{Bi-LSTM} & Max time step & 140 \\
		& Hidden neurons & 180 \\
		& Hidden layers & 2 \\
		& Batch normalization & True \\
		& Dropout rate & 0.5 \\ \hline
		\multirow{2}{*}{GCN} & Network size (size of $\Theta$) & {[}360,7{]} \\
		& Layers number & 1 \\ \hline
		Pooling & Pooling method & 50th percentile pooling \\ \hline
		\multirow{3}{*}{Regularization} 
		&Loss function & Cross entropy\\
		& L2 regularization & $1\times 10^{-8}$ \\
		& Orthogonal regularization & $1\times 10^{-8}$ \\ \hline
		\multirow{4}{*}{Training} & Optimizer & Adam \\
		& Mini-batch size & 32 \\
		& Learning rate & 0.001 \\
		& Weight decay rate & $1\times 10^{-8}$ \\ \hline
	\end{tabular}
\end{table}
\par
Fig. 5 and Fig. 6 show how the word embedding size and the number of hidden neurons in LSTM influence the F-measure of the model. These experiments show that the F-measure of our model is higher than 80\% with different word embedding sizes and different numbers of hidden neurons in LSTM. We fix the embedding size as 300 and number of hidden neurons in LSTM as 180 since they are likely to provide the best performance of our model. For the layers of GCN, we set it as 1. According to Marcheggiani and Titov \cite{Ref36}, when GCN is on the top of LSTM to deal with NLP task, single-layer GCN can best collaborate with LSTM. 
\begin{figure}[htbp]
	\centering
	\begin{minipage}[t]{0.48\textwidth}
		\centering
		\includegraphics[width=5.7cm]{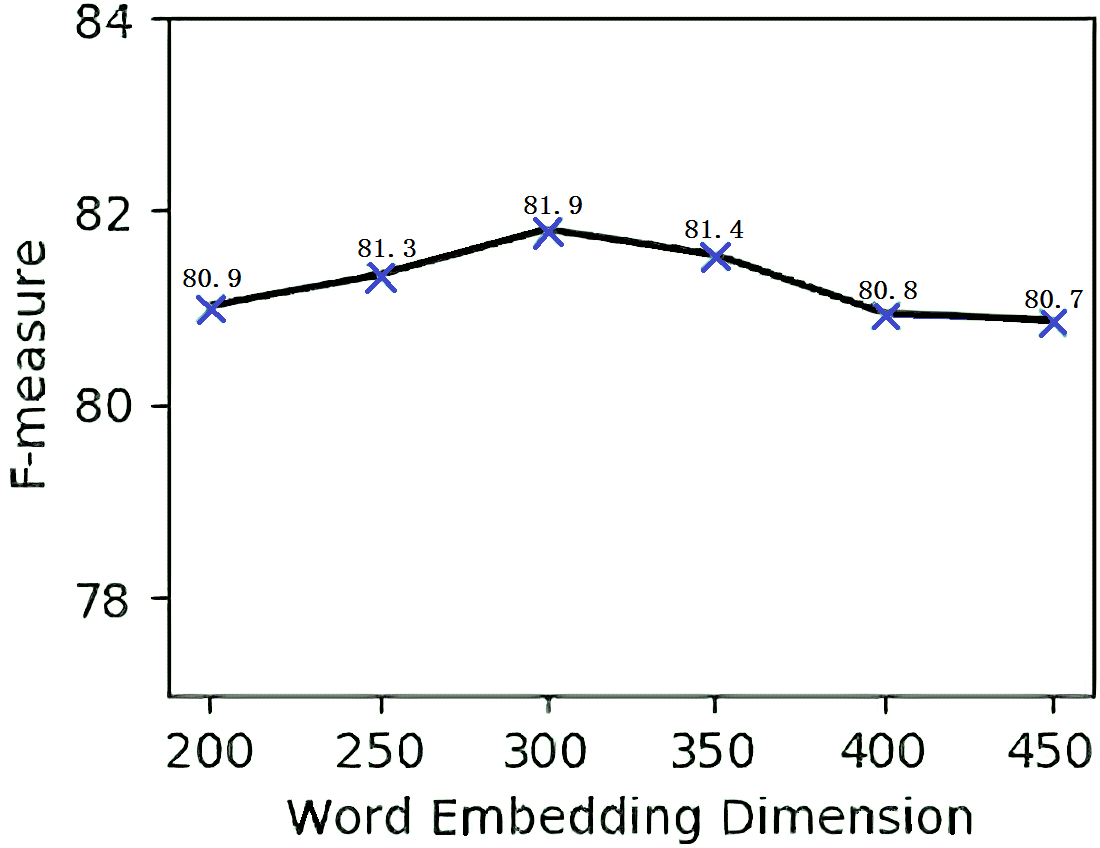}
		\caption{Word embedding size choosing experiment}
	\end{minipage}
	\begin{minipage}[t]{0.48\textwidth}
		\centering
		\includegraphics[width=5.7cm]{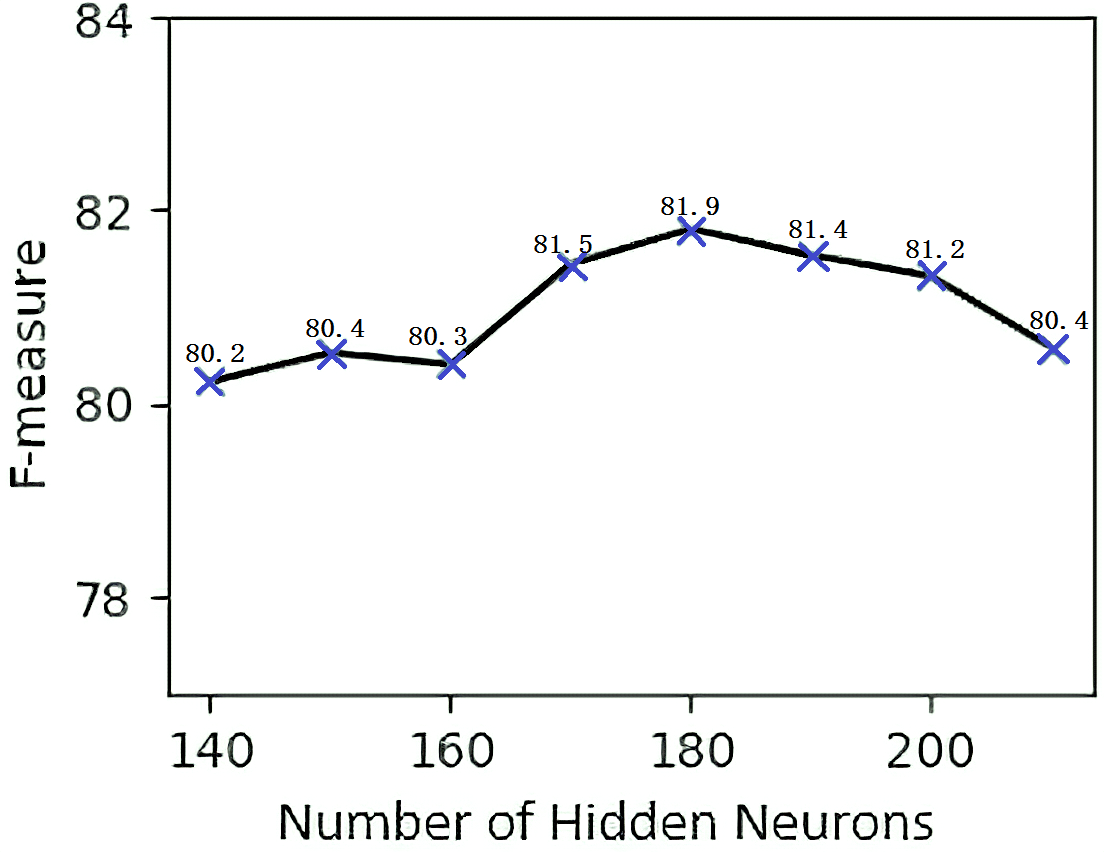}
		\caption{Number of hidden neurons choosing experiment}
	\end{minipage}
\end{figure}

\subsection{Comparison of Fine-grained Emotion Classification among Different Models}
We chose several sentiment classification algorithms as baselines, including both conventional machine learning methods and state-of-the-art neural network architectures. Results of our model against the best team in NLP\&CC2013 and other baselines are listed in Table 3.
\parskip=6pt
\par
\noindent\textbf{Benchmarks}
\begin{itemize}
\item[$\bullet$] CNN:
\par
In this approach, widely-used CNN model with one layer and max-pooling method is used. Other parts in the model are the same as others. 
\item[$\bullet$] LSTM:
\par
LSTM is also a powerful model in NLP tasks. Two-layer LSTM is employed to extract text features word by word. 

\item[$\bullet$] LSTM+CNN:
\par
Different from our model, this approach replaces GCN with CNN which is similar to GCN but do not have graph in it. This comparison shows both the contribution of GCN and syntax.
 
\item[$\bullet$] LSTM+GCN (without syntax information):
\par
In order to show the contribution of syntactic graph, we remove dependency parse tree in this experiment by assigning all element in the adjacency matrix with '1', so that every word links to others directly.

\end{itemize}

\begin{table}[htb]
	\caption{Comparison of different models on the NLP\&CC2013 testing dataset}
	\label{tab:3}       
	\begin{threeparttable}
		\begin{tabular}{lll}
			\hline
			Method & $Macro_{F-measure}$ & $Micro_{F-measure}$ \\ \hline
			Best team in NLP\&CC2013\tnote{$^{\#}$} & 31.29\% & 35.21\% \\
			Lexicon\tnote{$^{\#}$} & 30.77\% & 33.27\% \\
			SVM vote\tnote{$^{\#}$} & 23.62\% & 31.11\% \\
			CSRs\tnote{$^{\#}$} & 42.07\% & 44.19\% \\
			E-ESM (\cite{Ref38}) & 35.00\% & 43.90\% \\
			MCNN (\cite{Ref38}) & 34.62\% & 43.65\% \\
			EMCNN (\cite{Ref28}) & 35.17\% & 44.22\% \\
			CNN (\cite{Ref9})\tnote{*} & 67.42\% & 74.23\% \\
			LSTM (\cite{Ref39})\tnote{*} & 68.32\% & 73.42\% \\
			LSTM+CNN (replace GCN with CNN)* & 69.89\% & 76.42\% \\ \hline
			LSTM+GCN (without syntax information)* & 74.81\% & 77.99\% \\ 
			\textbf{Syntax-based GCN (Our Approach)} & \textbf{79.93\%} & \textbf{82.32\%} \\ \hline
		\end{tabular}
		\begin{tablenotes}\footnotesize
			\item[$^{\#}$] The result is cited from the paper of Wen \& Wan \cite{Ref22}.
			\item[*] It is conducted by ourselves on datasets shown in Table 1.
		\end{tablenotes}
	\end{threeparttable}
\end{table}
According to Table 3, our syntax-based GCN model outperformed the other models by 10.04\% on $Macro_{F-measure}$. Compared with the LSTM model and CNN network, LSTM with GCN attached could catch the context emotion information in the microblog more effectively, retaining the syntactic information of a sentence. Dependency parse tree is trained on a vast corpus using deep learning methods, which brings much information about language structure. 
\par
In general, as is shown in Table 3, syntax-based GCN with dependency parse tree increases both the $Macro_{F-measure}$ and $Micro_{F-measure}$ scores by around 5\% compared with LSTM + GCN without syntax information. Table 4 and Table 5 show the detailed categorical performance of our model with and without syntax information respectively. According to Table 4 and Table 5, syntax-based GCN network can make use of the sophisticated techniques of dependency parsing, especially for the emotion of 'Surprise' on which the syntactic information helps to increase the F1-score by 11\%. 

\begin{table}[htb]
	\caption{Detailed Result of Syntax-based GCN Emotion Detection}	\label{tab:4}  
	\begin{tabular}{llll}
		\hline
		& Precision & Recall & F1-score \\ \hline
		Happiness & 0.87 & 0.86 & 0.82 \\
		Sadness & 0.86 & 0.83 & 0.84 \\
		Like & 0.78 & 0.69 & 0.70 \\
		Anger & 0.88 & 0.85 & 0.87 \\
		Disgust & 0.82 & 0.86 & 0.84 \\
		Fear & 0.81 & 0.64 & 0.71 \\
		Surprise & 0.72 & 0.69 & 0.70 \\ \hline
		Micro Average & 0.82 & 0.82 & 0.82 \\
		Macro Average & 0.82 & 0.78 & 0.80 \\ \hline
	\end{tabular}
\end{table}

\begin{table}[htb]
\caption{Detailed result of LSTM+GCN(without syntax information)}
\label{tab:5}       
	\begin{tabular}{llll}
		\hline
		& Precision & Recall & F1-score \\ \hline
		Happiness & 0.76 & 0.81 & 0.79 \\
		Sadness & 0.86 & 0.73 & 0.79 \\
		Like & 0.78 & 0.81 & 0.79 \\
		Anger & 0.73 & 0.84 & 0.78 \\
		Disgust & 0.80 & 0.81 & 0.78 \\
		Fear & 0.74 & 0.66 & 0.70 \\
		Surprise & 0.70 & 0.51 & 0.59 \\ \hline
		Micro Average & 0.78 & 0.78 & 0.78 \\
		Macro Average & 0.77 & 0.74 & 0.75 \\ \hline
	\end{tabular}
\end{table}
\subsection{Comparison of different $p$th percentile pooling methods}
To test the validity of percentile pooling, we compare different $p$th percentile pooling methods with other widely used pooling methods. Comparison results are shown in Table 6.

\begin{table}[htb]
	\caption{Comparison of different $p$th percentile pooling}
	\label{tab:6}       
	\begin{tabular}{lll}
		\hline
		Pooling Method & $Macro_{F-measure}$ & $Micro_{F-measure}$ \\ \hline
		Fully Connected Layer & 71.25\% & 72.14\% \\
		30th percentile pooling & 75.87\% & 78.50\% \\
		40th percentile pooling & 69.75\% & 77.62\% \\
		50th percentile pooling & \textbf{79.93\%} & \textbf{82.32\%} \\
		60th percentile pooling & 74.05\% & 77.30\% \\
		70th percentile pooling & 77.20\% & 79.24\% \\
		80th percentile pooing & 76.19\% & 80.57\% \\
		Max (100th percentile) pooling & 74.02\% & 78.45\% \\
		Average Pooling & 74.75\% & 78.77\% \\ \hline
	\end{tabular}
\end{table}

According to Table 6, although the widely used max pooling works well, our models with 50th percentile pooling has the best performance, and that indicates percentile pooling fits syntax-based GCN model well. Experiments show that percentile pooling can improve the $Micro_{F-measure}$ of our model by 3.55\%. Besides, we tested 50th percentile pooling in the inferior CNN and LSTM models, which are showed in Table 7. It shows that 50th percentile pooling also outperformed max pooling by 1.63\% and 1.65\% respectively. 
\begin{table}[htbp]
	\caption{$Micro_{F-measure}$ of 50th percentile pooling and max pooling methods}
	\label{tab:7}       
	\begin{tabular}{lll}
		\hline
		Pooling methods & CNN(\cite{Ref9}) & LSTM(\cite{Ref39}) \\ \hline
		Max pooling & 57.58\% & 72.64\% \\ \hline
		50th Percentile Pooling & 59.21\% & 74.29\% \\ \hline
	\end{tabular}
\end{table}

\subsection{Comparison of different orthogonalization constraints}
We apply orthogonalization constraints penalty on both Bi-LSTM and GCN weight matrices to learn long-term dependence. The following experiments show how the coefficient parameter of the penalty influences the performance. Experiment results of different penalty coefficients are shown in Table 8. It shows, by setting the orthogonalization constraint to $1\times 10^{-8}$, our model can be effectively improved by 3.39\% and 5.54\% in $Micro_{F-measure}$ and $Macro_{F-measure}$ scores respectively.

\begin{table}[htbp]
	\caption{Comparison of different penalty coefficients}
	\label{tab:8}  
	\begin{tabular}{lll}
		\hline
		Penalty coefficient $\lambda$ & $Macro_{F-measure}$ & $Micro_{F-measure}$ \\ \hline
		$1\times 10^{-5}$ & 76.78\% & 79.60\% \\
		$1\times 10^{-6}$ & 76.37\% & 79.28\% \\
		$1\times 10^{-7}$ & 77.49\% & 80.66\% \\
		$1\times 10^{-8}$ & \textbf{79.93\%} & \textbf{82.32\%} \\
		0(no orthogonalization constraint) & 74.48\% & 78.93\% \\ \hline
	\end{tabular}
\end{table}

\subsection{Comparison of sentiment binary classification models}
Our model can be transformed into a polarity classification of positive and negative sentiments. In such case, microblogs with an ambiguous emotion type of \emph{surprise} are removed, and we consider emotion types of \emph{happiness}, \emph{like} belong to positive while \emph{anger, disgust, sadness,} and \emph{fear} belong to negative. To fit the binary classification problem, we set the output dimension of graph convolution network to 2, and other parts of the model remain unchanged. The comparison results of the different models are shown in Table 9.

\begin{table}[htbp]
	\caption{Comparison of polarity classification result}
	\label{tab:9}
	\begin{threeparttable}
		\begin{tabular}{llll}
			\hline
			Method & Accuracy & Precision & Recall \\ \hline
			Baidu AI open platform\tnote{$^{\#}$} & 76.49\% & - & - \\
			LSTM (\cite{Ref39})\tnote{*} & 90.01\% & 90.07\% & 90.15\% \\
			CNN (\cite{Ref9})\tnote{*} & 91.65\% & 91.62\% & \textbf{92.65\%} \\ \hline
			\textbf{Syntax-based GCN (Our Approach)} & \textbf{92.04\%} & \textbf{92.19\%} & 92.04\% \\ \hline
		\end{tabular}
		\begin{tablenotes}\footnotesize
			\item[$^{\#}$] https://ai.baidu.com/tech/nlp/sentiment\_classify.
			\item[*] It is conducted by ourselves on datasets shown in Table 1.
		\end{tablenotes}
	\end{threeparttable}
\end{table}

According to Table 9, the syntax-based graph convolution network (GCN) model also performs well in polarity sentiment classification with higher accuracy and precision. Experiments show that syntax-based GCN improve the performance slightly in binary emotion classification task. One possible reason for this result is that binary classification task is relatively simple, and its accuracy is already high, so every 1\% improvement can be very challenging, while in fine-grained task, it is more complex, for which more information contributed by GCN with dependency parse tree is useful.

\section{Conclusions and Future Work}
In this paper, we introduced a syntax-based graph convolution network (GCN) model for fine-grained emotion detection. To the best of our knowledge, this is the first work to integrate syntactic information in the GCN for emotion detection. Experiments show that our model has fully exploited the syntactic relation of a sentence, which outperforms state-of-the-art algorithms. Besides, we proposed a new feature pooling method for neural networks which promotes the robustness of the model. Finally, we contribute a new annotated dataset of Chinese microblogs to public research.
\par
It is an interpretable phenomenon that syntactic information boosts the performance of our model. However, the method we used to construct sentence graph from word wise dependence parsing is still primitive, declaring a grander prospect of our model. In this paper, exact types of syntactic dependency relationship are ignored, and the choosing of $p$th in percentile pooling is arduous. Future work can incorporate more information about syntactic dependency relationship. Besides, the $p$th percentile pooling can be designed to be self-adapted during training or try k-percentile pooling that similar to the k-max pooling method. Last but not least, in this paper, we only focus on Chinese microblogs. When we apply our model to English corpus (Stanford Sentiment Treebank), the model does not show its advantages compared to the state-of-the-art. Future work can be done on how to adapt the model to other languages such as English. 
\section{Acknowledgments}
This work is supported by the National Key R\&D Program of China (No. 20-16YFC1401900), the National Natural Science Foundation of China (61173029, 61672144, 61872072), and the Australian Research Council Discovery Grants (DP170104747, DP180100212).



\end{document}